# Development of Low-Cost IoT Units for Thermal Comfort Measurement and AC Energy Consumption Prediction System

*Yutong* CHEN[1*], *Daisuke* SUMIYOSHI[2], *Riki* SAKAI[1], *Takahiro* YAMAMOTO[3], *Takahiro* UENO[4], and *Jewon* OH[5]

[1] Kyushu University, Graduate School of Human-Environment Studies, Japan
[2] Kyushu University, Faculty of Human-Environment Studies, Japan
[3] Kagawa University, Faculty of Engineering and Design, Japan
[4] Waseda University, Faculty of Science and Engineering, School of Creative Science and Engineering, Japan
[5] Sojo University, Faculty of Engineering Department of Architecture, Japan

**Abstract.** In response to the substantial energy consumption in buildings, the Japanese government initiated the BI-Tech (Behavioral Insights X Technology) project in 2019, aimed at promoting voluntary energy-saving behaviors through the utilization of AI and IoT technologies. Our study aimed at small and medium-sized office buildings introduces a cost-effective IoT-based BI-Tech system, utilizing the Raspberry Pi 4B+ platform for real-time monitoring of indoor thermal conditions and air conditioner (AC) set-point temperature. Employing machine learning and image recognition, the system analyzes data to calculate the PMV index and predict energy consumption changes due to temperature adjustments. The integration of mobile and desktop applications conveys this information to users, encouraging energy-efficient behavior modifications. The machine learning model achieved with an $R^2$ value of 97%, demonstrating the system's efficiency in promoting energy-saving habits among users.

## 1 Introduction

Achieving carbon neutrality society efforts related to buildings are important, which account for a large share of $CO_2$ emissions. The earliest research on energy efficiency in buildings is 1974, when V.L. Sailor [1] analyzed building energy efficiency calculation models in commercial buildings. In the decades since, more and more scholars have been working on the field of energy efficiency in buildings. Clyde Zhengdao Li [2] analyzed 2569 articles on building energy efficiency from 1974 to 2020, with residential buildings as a keyword in a large proportion of them.

However, unlike residential buildings, energy efficiency in office buildings is more difficult, especially for small and medium-sized office buildings which have less funding and energy strategies. In addition, there is little research focusing on these kinds of buildings (Fig. 1).

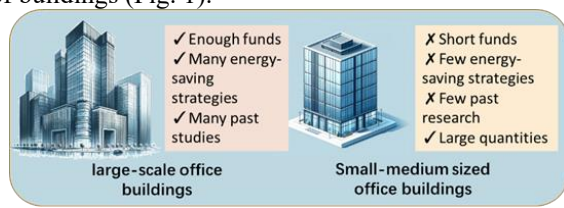

**Fig. 1.** Difference in small-medium sized office buildings

Under this background, the Ministry of the Environment of Japan proposed the BI-Tech (Behavioral Insights X Technology) project aiming to promote voluntary behavior change through the combined of behavioral insights and Artificial Intelligence (AI) & Internet of Things (IoT) technologies. Facing the character of small and medium-sized office buildings lack of funds and energy-saving control strategies, cultivating occupants' behavioral energy-saving awareness is becoming potential research.

This study presents the development of a BI-Tech application platform, designed for small and medium-sized office buildings, which integrates thermal comfort and energy consumption prediction. By leveraging real-time indoor thermal comfort data, obtained from IoT devices, and predicting changes in air conditioner (AC) energy consumption in response to 1°C temperature adjustments by occupants, the platform aims to promote optimal AC usage.

## 2 Related works

### 2.1 Thermal comfort assessment system

The American Society of Heating, Refrigerating and Air-Conditioning Engineers (ASHRAE) defined thermal comfort as "the condition of the mind in which satisfaction is expressed with the thermal environment." In the evaluation of thermal comfort models, Fanger's PMV (Predicted Mean Vote) - PPD (Predicted Percentage Dissatisfaction) thermal comfort model has made a pioneering contribution to thermal comfort theory and the assessment of indoor thermal environments in buildings. It has been widely used for thermal comfort design and site assessment [3]. However, the acquisition of physical quantities such as mean radiant temperature ($\bar{t}_r$) and air velocity ($V_{air}$) included in the PMV index is challenging, and the calculation process requires the operation of multiple higher-order equations. Therefore, an increasing number of related studies are choosing to use methods such as machine learning (ML) to predict instead of measurement.

* Corresponding author: chen.yutong.262@s.kyushu-u.ac.jp

Luo [4], compared the performance of different ML algorithms and shows that factors such as building type, building operation mode and climatic conditions although not the primary influences on human thermal comfort affect how people perceive heat. Ngarambe [5] reviewed 37 papers from 2005 to 2019 to investigate AI-based thermal comfort prediction models, the energy impact of AI-based thermal comfort control, machine learning methods and algorithms for thermal comfort models, PMV models, and Personal Comfort Models (PCM) models.

While methods exist to obtain Predicted Mean Vote (PMV) values, ensuring their timeliness and accuracy is crucial for thermal comfort. Machine learning can predict PMV data, but it is not without errors compared to actual calculations. Furthermore, the process of training and generalizing the model is time-consuming. Addressing the challenge of real-time collection of the six factors affecting PMV and calculating the PMV index in real-time is imperative.

## 2.2 IoT using in thermal comfort assessment

As buildings become smarter through AI, networks, and IoT, new opportunities are given to address greater challenges in building operations, design, and user experience [6]. The application of IoT technology makes data collection easier, especially in collecting data like mean radiant temperature and air velocity which were difficult to collect in the past. These parameters are of significant importance in the calculation of PMV.

Farid Ali Mousa [7], proposed a low-cost measurement and predicted PMV value through the Chicken Swarm Optimization (CSO) algorithm that reached 98.3% accuracy. However, this study didn't detect mean radiant temperature which have the great influence of PMV value.

Giacomo Chiesa [8] employed Raspberry Pi and Arduino technologies to develop a platform for collecting and evaluating thermal comfort indicators PMV and PPD, offering a practical application for occupants to monitor their thermal environment. Mehrzad Shahinmoghadam [9] introduced a virtual reality (VR) tool aimed at the real-time assessment of thermal comfort through the integration of Building Information Modeling (BIM) and Internet of Things (IoT) data. Similarly, to Giacomo Chiesa, providing only numerical feedback does not significantly improve user interaction, especially for those inexperienced with such data. This approach fails to fully engage users or help them understand thermal comfort to enhance their lifestyle. Despite considerable research into IoT for

| Nomenclature | |
|---|---|
| M | Metabolic rate (W/m$^2$) |
| W | Effective mechanical (W/m$^2$). |
| $I_{cl}$ | Clothing insulation (m$^2$ K/W). |
| $f_{cl}$ | Clothing surface area factor |
| $t_a$ | Air temperature (°C) |
| $\bar{t}_r$ | Mean Radiant Temperature (°C) |
| $V_{air}$ | Relative air velocity (m/s). |
| Pa | Water vapor partial pressure (Pa) |
| $h_c$ | Convective heat transfer coefficient (W/m$^2$K) |
| $t_{cl}$ | Clothing surface temperature (°C) |
| $t_g$ | Globe temperature (°C) |
| Tout | Outdoor temperature (°C) |
| W | AC energy consumption (kWh/win) |
| IoT | Internet of Things |
| AC | Air Conditioner |
| PMV | Predicted Mean Vote |
| PPD | Predicted Percentage of Dissatisfied |
| ML | Machine Learning |
| SVM | Support Vector Machine |
| SVC | Support Vector Classifier |
| ANN | Artificial neutron network |
| DT | Decision Tree |
| RF | Random Forest |
| SVR | Support Vector Regression |

thermal comfort monitoring, effectively communicating this data to users remains an unresolved challenge.

## 2.3 Prediction of building energy consumption

As for a type of black box model, ML acquire limited detailed physical characteristics of the buildings but still could predict the energy consumption.

Aydinalp [10] used ANN to estimate the heating and cooling loads of a building. Wong [11] used ANN to evaluate the dynamic energy performance of a commercial building with daylighting in Hong Kong.

Li [12] applied SVM to predict monthly and short-term (daily) electricity consumption forecasts for a residential building located in Japan.

RF (Radom Forest) is a model composed of many decision trees, commonly used for both classification and regression problems. In many instances, RF models have been proven to possess excellent predictive capabilities.

Wang [13] introduced an hourly building energy consumption prediction method based on RF. They compared RF with Regression Trees (RT) and Support Vector Regression (SVR). According to the performance index (PI), the RF prediction performance was found to be better by 14%-25% compared to RT and 5%-5.5% compared to SVR.

Past studies present the potential for using ML in prediction tasks. However, some of these studies primarily rely on historical data, lacking the capability for automatic updates based on real-time data. This limitation hinders improvements in prediction accuracy. Additionally, the concepts of energy conservation and thermal comfort are in conflict. Discussion on how to effectively balance these two aspects is essential for providing occupants with strategies to attach energy-saving behavior.

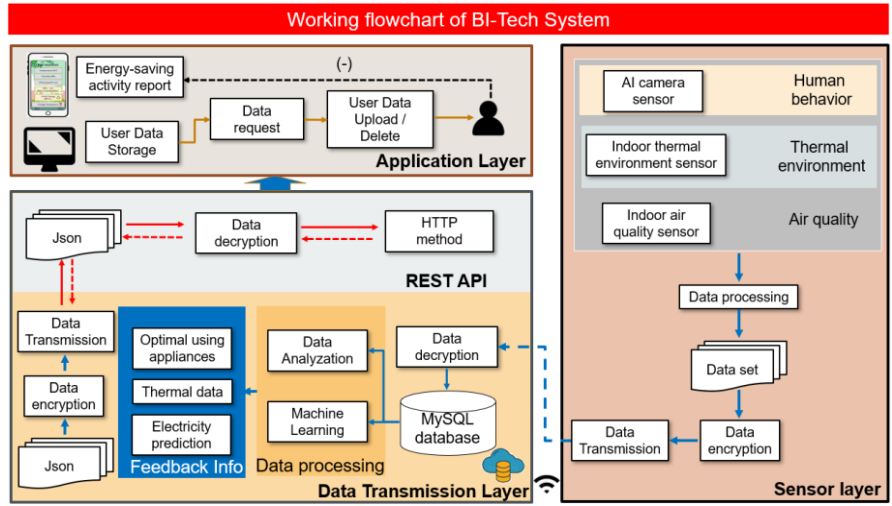

**Fig. 2.** Architecture of BI-Tech system

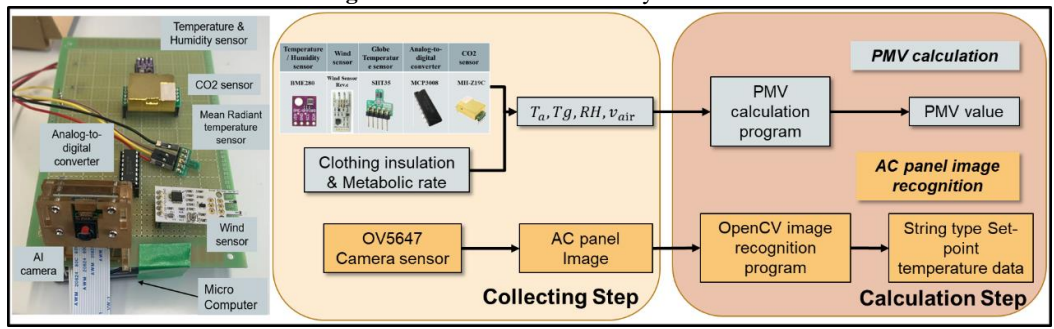

**Fig. 3.** Measurement steps of IoT system

### 2.4 Work on this study

When setting the temperature of an AC, users may often do not have a clear concept of what the appropriate setting should be. Sometimes, a mere 1°C adjustment in the AC's settings may have a negligible impact on thermal comfort. However, this slight alteration can lead to substantial variations in energy usage.

Based on past studies, this study introduced a BI-Tech system, offering a new approach to integrating the concept of thermal comfort with energy consumption prediction. This study employs IoT modules for live thermal information collection and PMV calculation, uses ML to predict AC energy consumption based on real-time data. Providing users with feedback information via application platforms on the predicted changes in energy consumption and $CO_2$ emissions for each 1°C adjustment in temperature.

The integration of thermal comfort and energy consumption, communicated with occupants through the app, not only enhances occupants' understanding of thermal comfort but also has the potential to guide users in changing AC settings to foster energy-saving behaviors while limited change of thermal comfort.

- A low-cost indoor thermal comfort measurement and AC set-point temperature image recognition system has been developed.
- An SVM-RF machine learning AC energy consumption prediction model has been built to predict energy change when occupants turn up/down 1°C set-point temperature based on real-time data.
- An iOS system application and Desktop application system have been developed to provide feedback for occupants' measurement data and energy-saving strategies generated by a notification system.

## 3 Structure of BI-Tech system

The BI-Tech system integrates IoT, ML, and App modules. An IoT module has been used to monitor real-time data and calculate thermal comfort according to ASHRAE 55 standards through the PMV-PPD index. In summer, the system employs ML to forecast changes in AC energy use and $CO_2$ emissions for every 1°C-temperature adjustment when set-point temperature above or below recommended value based on current data. Users receive this environmental information and predictions via the app. The merge of thermal comfort and AC energy consumption prediction has provided the potential to guide users towards energy-saving actions.

As shows in Fig. 2, the BI-Tech system proposed in this study is divided into three main layers. They are the sensor layer, the data transmission layer and the user interaction layer based on the iOS system and Windows system. These three layers are connected to each other in a progressive manner to achieve the final effect of interaction with the user.

The sensor layer mainly collects core data for the system and performs data preprocessing. In this layer, the core data based on thermal comfort calculation such as mean radiant temperature and air velocity are collected and pre-processed to get the value of thermal comfort PMV index. At the same time, through the camera, the set-point temperature of the AC panel is recognized and converted into an array of string type. These data are processed and passed to the data transfer layer for secondary processing and ML model training through Message Queue Telemetry Transport (MQTT) communication transfer protocol.

The data transmission layer primarily receives preprocessed data from sensor layer, performs secondary processing and storage of this data, and facilitates the transfer of training data for machine learning models to the application layer. Data collected by the sensor layer, after undergoing initial preprocessing, is transmitted to cloud-based servers using the MQTT communication protocol. Once on the cloud, the data is stored in a MySQL database and undergoes secondary processing. This secondary processing includes handling the SVM-RF ML dataset, training predictive models, and generating output results. The outcomes are then conveyed to the application layer in JSON format via REST (Representational State Transfer) API.

The application layer of this system, comprising an iOS and a Windows desktop application, serves as an intermediary between users and back-end components. Unlike traditional dashboards, this study employs a combination of a notification systems to provide users with real-time feedback on energy-saving strategies derived from back-end processing, thereby guiding them towards adopting energy-saving behaviors.

## 4 Back-end of BI-Tech system.

**4.1 Hardware kit.**

In this study, five sensor modules and one digital-to-analog converter module based on Raspberry pi 4B+ were selected to measure the indoor air temperature, globe temperature, relative humidity, air velocity and set-point temperature data of the AC panel. (Table 2). The measurement process has two steps (Fig. 3). Through the selected sensor physical quantities and image information captured by the sensor module after processing the output of the thermal comfort PMV value and the AC set-point temperature in string type by Python program. The dataset was constructed for the next step of building the AC energy consumption prediction model based on AC set-point temperature.

Table 2. Selected modules

| Items | Sensor Model | USD | Accuracy |
|---|---|---|---|
| $t_g$ | SHT35 | 10.22 | ± 0.1°C |
| $t_a$ & Rh | BME280 | 6.78 | ± 0.5°C (at 25°C) ± 3% RH |
| $V_{air}$ | Wind Sensor Rev. C | 22.39 | ± 0.01 m/s |
| ADC | MCP3008 | 3.07 | 10 bits |
| $CO_2$ | MH-Z19C | 10.56 | ± 50 ppm |
| Camera | OV5647 | 4.74 | 1080 p |

*4.1.1 Thermal comfort assessment.*

The PMV model considers the influence of external environmental factors ( $t_a, \bar{t}_r, h_a, V_{air}$ ) and two non-environmental factors ( M, $i_{cl}$ ) on human comfort. Factors such as Mean Radiant Temperature ($\bar{t}_r$) and other physical quantities cannot be directly measured using sensors, making the direct calculation of the PMV value challenging. This is due to the highly nonlinear and iterative nature of the formulas associated with it [3].

This study measured $t_g$ using equation (1) to calculate $\bar{t}_r$, the SHT35 temperature sensor was placed in a 150 mm diameter black copper ball to measure the indoor globe temperature. In the case of the standard globe D=150 mm,

$$\bar{t}_r = [(t_g + 273)^4 + 0.4 \times 10^8 |t_g - t_a|^{\frac{1}{4}} \times (t_g - t_a)]^{\frac{1}{4}} - 273 \quad (1)$$

Physical quantities such as (indoor air temperature, black globe temperature, relative humidity, air velocity) are collected and pre-processed by the sensors and combined with the clothing insulation selected according to the seasons and the metabolic rate daily to form the dataset.

To address the nonlinear and iterative equations for PMV value calculation, this study employs Sympy and Math libraries in Python. Equations are bifurcated into higher order and general equations for computation (Fig. 4) Complex numbers and numerical outliers arising from the quartic equation solution are methodically filtered out using Python's conditional directives.

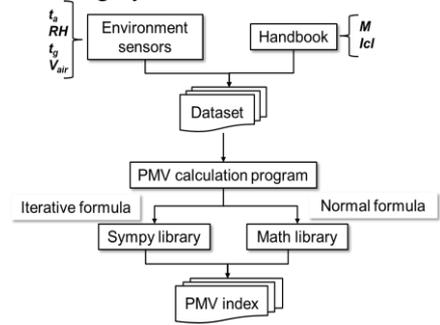

**Fig. 4.** Calculation flow of PMV value

*4.1.2 Set-point temperature image recognition.*

In office buildings, AC is often controlled by wall panels, making direct temperature setting access difficult without changing the AC system. This research employs a camera sensor module and OpenCV technology for image recognition to identify the set-point temperature the process is mainly divided into the steps of image contrast enhancement, threshold processing, grayscale processing, erode processing contour extraction and final comparison with the sample data (Fig. 5).

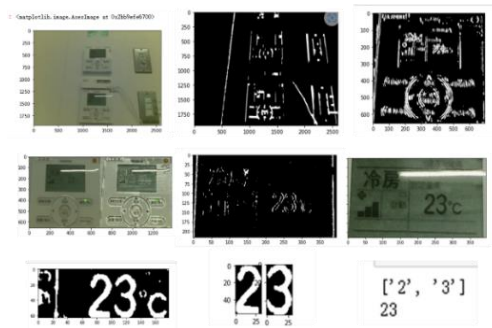

**Fig. 5.** Set-point temperature image recognition

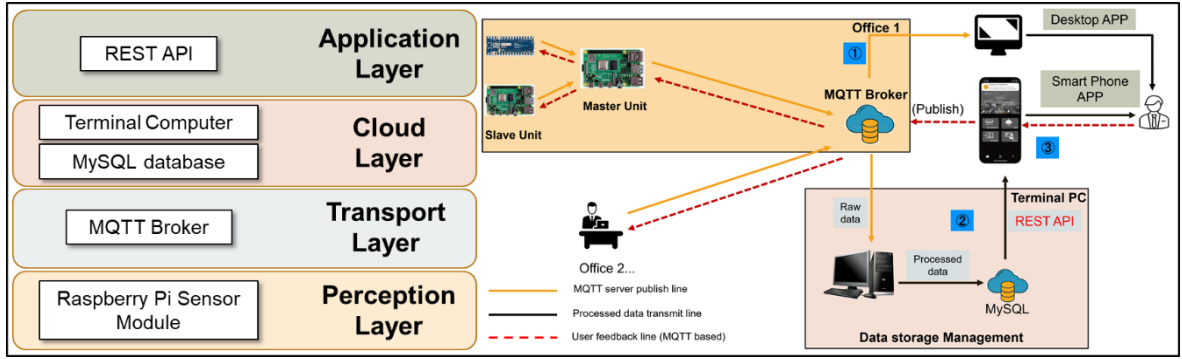

**Fig. 6.** Data transmission structure

### 4.2 Cloud & database

This study adopts the MQTT communication protocol as the fundamental framework for data transmission throughout the system. Employing a Publish/Subscribe model, MQTT facilitates the interconnection of diverse IoT devices, allowing for seamless communication via clients subscribing to various topics. Within this paradigm, a message published by one client on a specific topic is conveyed to another client subscribed to that topic via the MQTT broker. This architecture distinguishes MQTT from the conventional Request/Response model of HTTP, rendering it more adept at supporting real-time data exchange and ensuring swift reconnection after disruptions. Furthermore, MQTT's Quality of Service (QoS) mechanism is critical for maintaining the reliability of message delivery, effectively preventing data loss.

We independently developed a backend cloud system utilizing the MQTT communication protocol combined with a REST API. As depicted in the architecture, data is measured by Raspberry Pi sensor modules at the perception layer and transmitted to the MQTT Broker via the MQTT protocol. This data is then received by a terminal computer, stored in a MySQL database for further processing. After undergoing a series of ML training, the REST API serves as an intermediary for data exchange and interaction between the database and desktop or smartphone applications. Throughout this information exchange, multi-layer encryption is employed to ensure data security (Fig. 6).

### 4.3 AC energy prediction mode

This study developed an AC energy consumption prediction model based on indoor and outdoor thermal environment information, along with AC set-point temperature and energy consumption data, through four steps. The first step involves data collection and pre-processing through IoT modules, followed by correlation analysis and dataset pre-training. The second step entails model evaluation. Initially, a classification model is selected, and the SVM model is used to analyze the dataset's features, eliminating less significant ones to enhance the predictive accuracy of the regression model. The processed data is then inputted into regression models such as RF, ANN, LR, SVR for cross-validation to select the model with the highest accuracy. The third step involves tuning the selected predictive model using the Grid-search method until the most accurate parameters are found for AC energy consumption prediction based on set temperature. The fourth step, in the practical application of the BI-Tech system, combines real-time physical data for energy consumption prediction under actual environmental conditions (Fig. 7).

#### 4.3.1 Dataset processing

This study conducted ML data collection in the student room at Kyushu University's School of Human Environment Studies. Data was collected during the summer months (August-September) of 2022 and 2023. The arrangement of the measurement devices is illustrated in the Fig.8, considering the impact of different areas on environmental monitoring.

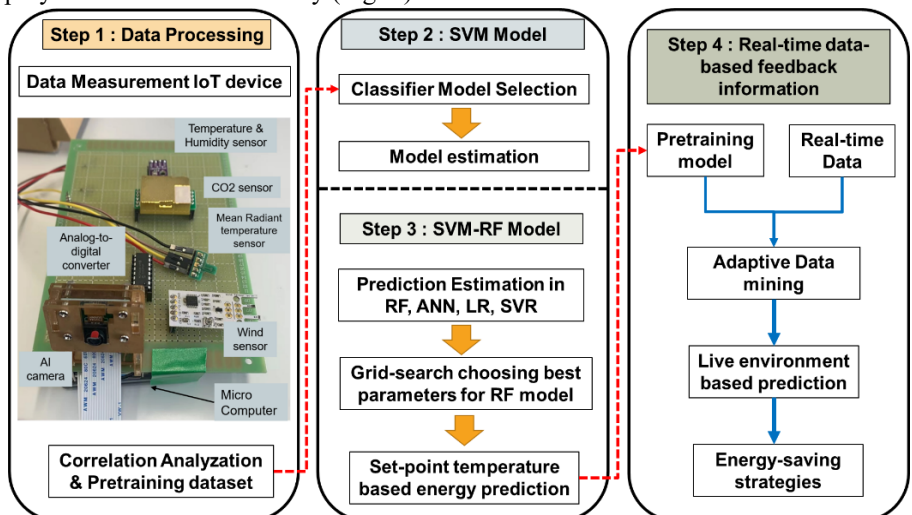

**Fig. 7.** Building steps of SVM-RF model

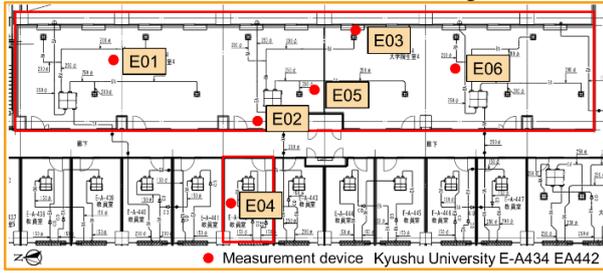

**Fig. 8.** Arrangement of the measurement devices

The Fig. 9 shows PMV and MRT data recorded during 2023 August 25th to 31st at five measurement points and outdoor temperature. As indicated in the device layout diagram, Measurement Point 1 is located below the AC outlet, resulting in consistently lower PMV values compared to other points. Measurement Point 4, affected by the western sun in summer, shows higher PMV values. Due to the same reason, the MRT fluctuation at Point 4 is the largest among all points. Both charts demonstrate that the trend in MRT changes closely mirrors that of PMV.

The dataset in this study is derived from three sources. The data of outdoor weather conditions are obtained from the official website of the Japan Meteorological Agency. Indoor thermal environment data and AC set-point temperature data were obtained from real-time data collection from the IoT system. The AC energy consumption data during the experiment period of the system was obtained from the power sensors set up by the administration of Kyushu University.

Fig. 10 presents the correlation analysis among various variables. The study underscores the substantial impact of set-point temperature alterations on energy consumption, highlighting the potential for energy savings through strategic adjustments. Specifically, a notable negative correlation (r = -0.26) is observed between changes in set-point temperature and AC energy usage, emphasizing the effectiveness of set-point modulation in energy conservation strategies.

Fig. 11 illustrates the histogram of the frequency distribution of AC energy consumption per minute in cooling mode. The distribution of energy consumption is relatively concentrated, indicating that most of the energy values are concentrated in certain intervals. The data at the tail of the histogram reveals the presence of extreme values or outliers in the data. Using distinct regression predictions for various frequency intervals enables the model to identify and forecast patterns in both high and low frequency ranges more precisely. This approach improves the model's ability to process the bulk of the data effectively and to predict values that are crucial for its overall predictive performance. In order to reflect different patterns of energy consumption behavior, this study first uses a classifier model to discretize the energy consumption data to train a regression model against a specific energy interval distribution, thus achieving higher accuracy in continuous prediction.

This preparatory step ensures that the regression model is provided with detailed, nuanced data that captures potential changes and trends in energy use.

### 4.3.2 Estimation of classifier model

In selecting classifiers, we compared learning curves of 3 different models. Using a dataset of 2500 as a baseline, contrasting the training and testing datasets revealed three distinct learning curves. As Fig. 12 shows the Gradient Boosting Classifier model and KNeighborsClassifier model has the lower scores, the SVM model has the best accuracy demonstrates greater potential.

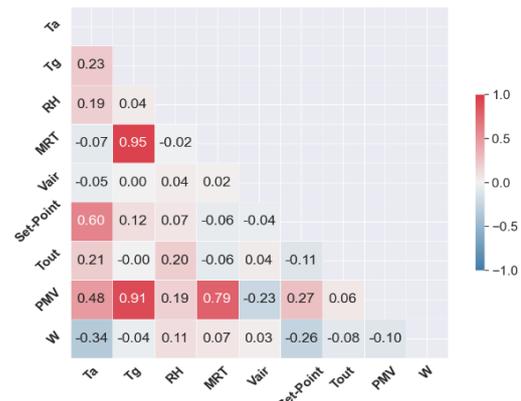

**Fig. 10.** Correlation heatmap matrix of input features

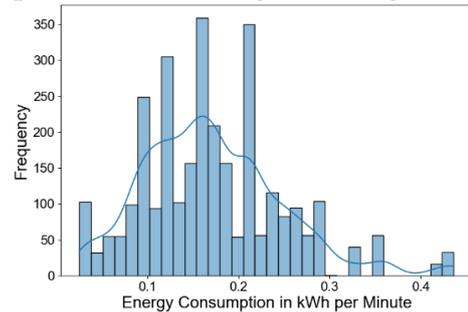

**Fig. 11.** Distribution of AC energy consumption per minute

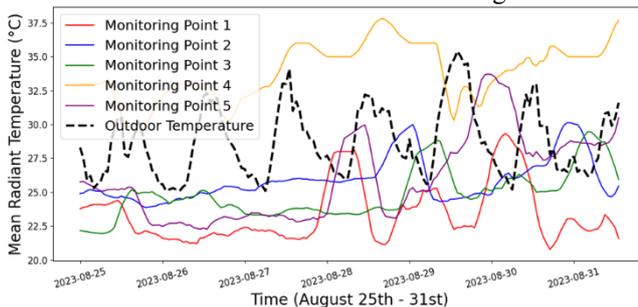
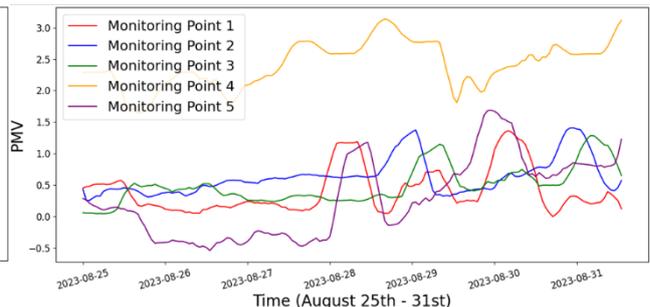

**Fig. 9.** PMV and MRT data from August 25th to 31st

### 4.3.3 Cross-validation of SVM Model

In the SVM model, cross-validation plays a crucial role in determining the optimal parameters C and Gamma, thereby enhancing the accuracy of the model's predictions. In this study, we choose Grid Search method to find the best C and Gamma (Table 3).

**Table 3.** Grid-search result

| Parameters | Setting of Numpy.linspace | | | Value | Score |
|---|---|---|---|---|---|
| | Start | Stop | Interval | | |
| C | 0.01 | 100 | 50 | 8.172 | 0.83 |
| Gamma | 0.1 | 30 | 50 | 0.710 | 0.82 |

### 4.3.4 Estimation of SVM-RF model

We compared the predictive performance of four models. After inputting two hundred validation sets, the Random Forest model exhibited the highest fit and the best performance. An $R^2$ value of 97% indicates that the model has excellent predictive ability.

As shows in Fig. 13, the comparsion of prediction results with actual data of the SVM-RF model, SVR, ANN and linear regression. SVM-RF model is the closest to the actual values (Red mark). This indicates a high degree of overlap between the predicted values and the actual values, suggesting high model accuracy.

### 4.3.5 Comparison of Loss Functions

Root Mean Squared Error (RMSE), Mean Squared Error (MSE), and Mean Absolute Error (MAE) are three crucial metrics for evaluating the performance of a model. As Fig. 14 illustrates, the RMSE, MSE, and MAE values of the SVM-RF model are all within a relatively small range. Comparing to SVR, ANN and Linear Regression, random forest has the lowest loss in model, indicating its high accuracy.

## 5 Front-end of BI-Tech system

In this study, the role of the desktop and iOS applications is to serve as platforms that communicate the thermal comfort PMV values obtained from IoT measurements, and the energy consumption values predicted through machine learning to users.

### 5.1.1 Function of data acquisition

By accessing data processed via REST API and the MQTT Broker, occupants are enabled to retrieve current thermal information, and $CO_2$ levels. Furthermore, through interaction with the cloud database using REST API, users can utilize the desktop application to track and analyze the variations in these indoor environmental indicators over the preceding 24 hours.

### 5.1.2 Notification system

Simply displaying raw measurement data is insufficient for providing users with actionable insights. An integrated approach, combining these measurements with a notification system, presents a more viable solution. In this study, the system is configured to issue alerts when air conditioning (AC) settings deviate from established optimal temperature thresholds—set below 26°C during summer months and above 25°C throughout winter.

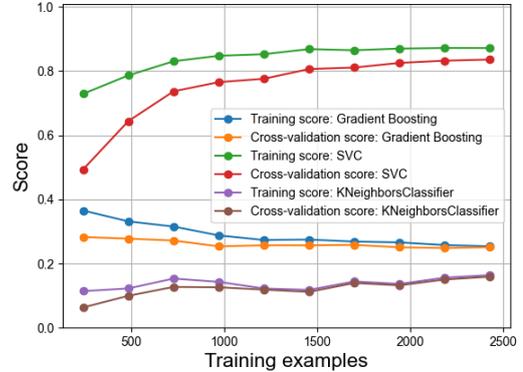

**Fig. 12.** Learning curve of 3 different classifier models

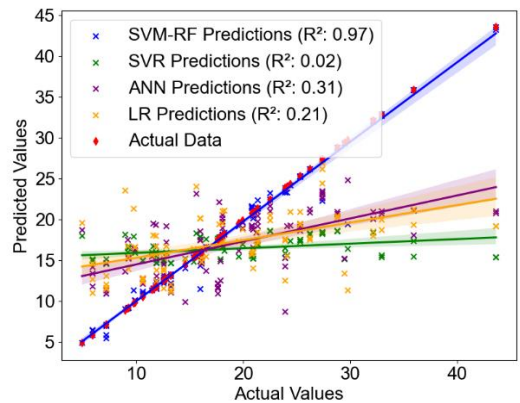

**Fig. 13.** Comparison of prediction results with actual data of SVM-RF, SVR, ANN and Linear Regression

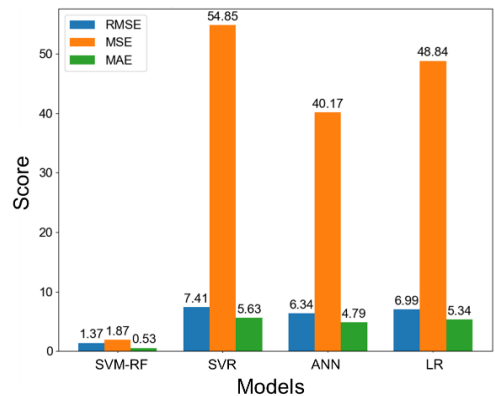

**Fig. 14.** RMSE, MSE, MAE comparison of SVM-RF, SVR, ANN, Linear Regression

Upon detection, the system advises users to make a minor adjustment of 1°C to the current temperature setting, subsequently providing them with feedback on the potential energy savings and reduction in $CO_2$ emissions that could be achieved. Moreover, should the indoor temperature veer outside the recommended parameters, the system proactively engages users with suggestions aimed at harmonizing energy efficiency with thermal comfort. Adhering to these guidelines allows users to participate in a points accumulation program, designed to reward energy-conserving practices. User actions are documented and evaluated within a cloud-based leaderboard, cultivating a competitive yet cooperative atmosphere that encourages ongoing commitment to energy conservation.

Fig. 15 shows desktop application interfaces, for viewing data from the past 24 hours from MySQL database, real-time data checking.

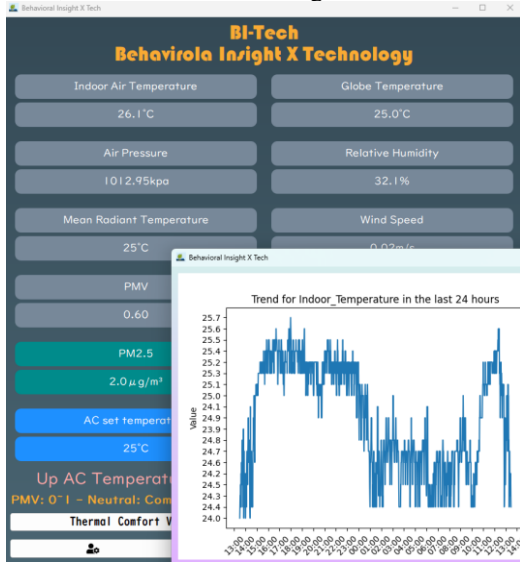

**Fig. 15.** Desktop application interface

Fig. 16 shows iOS application interfaces, including real-time data checking, notification interface and predicting result of AC energy consumption and $CO_2$ emmission while decrease 1 °C of set-point temperature in winter. Notification system combined thermal comfort and AC energy precition, aiming to guide the occupants saving energy while keep thermal comfort.

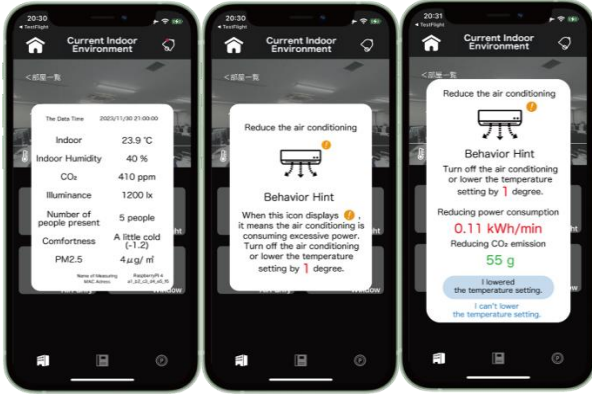

**Fig. 16.** iOS application interface

## 6 Discussion

This study proposes a novel approach for small and medium-sized buildings, addressing the balance between real-time thermal comfort monitoring and AC energy consumption prediction.

The applications developed for this study facilitate real-time user interaction, effectively disseminating processed environmental data. The IoT module is adept at gathering continuous indoor thermal environmental metrics, calculating the PMV index, and recognizing the AC set-point temperature. The ML model developed for predicting AC energy consumption, based on set-point temperature has demonstrated robust predictive capabilities, evidenced by an $R^2$ value of 97% following comprehensive model evaluation and cross-validation. This study presents BI-Tech system considerable potential for promoting energy conservation in small and medium-sized buildings.

There are many factors that this study has not yet considered, and future improvements will focus on the following areas :

- Redesigning IoT devices using PCB technology to create a more compact and convenient system.
- Enhancing the monitoring of clothing insulation using image recognition technology to improve the accuracy of the PMV index.
- Developing strategies for the spring and autumn seasons to discourage premature use of AC system. When outdoor air quality and thermal conditions are favorable, recommending natural ventilation to users as an alternative to mechanical ventilation, thus reducing energy consumption.
- Conducting experimental validations of the applications proposed by the BI-Tech system to enhance and improve the system's energy efficiency.

## Acknowledgement

This work was supported by JST, the establishment of university fellowships towards the creation of science technology innovation, Grant Number JPMJFS2132.